\documentclass[runningheads]{llncs}
\usepackage{graphicx}
\usepackage{amsmath,amssymb} % define this before the line numbering.
\usepackage{color}
\usepackage[pagebackref=true,breaklinks=true,letterpaper=true,colorlinks,bookmarks=false]{hyperref}

\usepackage[width=122mm,left=12mm,paperwidth=146mm,height=193mm,top=12mm,paperheight=217mm]{geometry}

\usepackage{epsfig}
\usepackage{graphicx}
\usepackage{stmaryrd}
\usepackage{algorithm}
\usepackage{algorithmic}
\usepackage{subfigure}
\usepackage{tabularx}
\usepackage{tabularx}
\usepackage{multirow}
\usepackage{array}
\usepackage{enumerate}
\usepackage{paralist}
\usepackage{tabularx}

\newcommand\etal{\emph{et.al.}} 
\newcommand\ie{\emph{i.e.}} 
\newcommand\cf{\emph{c.f.}}

\begin{document}
% \linenumbers
\pagestyle{headings}
\mainmatter

\title{\large Stereo Computation for a Single Mixture Image}
% Replace with your title

\titlerunning{Stereo Computation for a Single Mixture Image}
% Replace with a meaningful short version of your title

\authorrunning{Y. Zhong and Y. Dai and H. Li}
% Replace with shorter version of the author list. If there are more authors than fits a line, please use A. Author et al.

\author{Yiran Zhong\inst{1,3,4}\and
Yuchao Dai\inst{2} \and
Hongdong Li\inst{1,4}}
%% First names are abbreviated in the running head.
\institute{Australian National University \and
Northwestern Polytechnical University \\ \and
Data61 CSIRO  \and
Australian Centre for Robotic Vision \\
\email{\{yiran.zhong,hongdong.li\}@anu.edu.au \
daiyuchao@nwpu.edu.cn}}

\maketitle

\begin{abstract}
This paper proposes an original problem of \emph{stereo computation from a single mixture image}-- a challenging problem that had not been researched before.  The goal is to separate (\ie, unmix) a single mixture image into two constitute image layers, such that the two layers form a left-right stereo image pair, from which a valid disparity map can be recovered. This is a severely illposed problem,  from one input image one effectively aims to recover three (\ie, left image, right image and a disparity map).  In this work we give a novel deep-learning based solution, by jointly solving the two subtasks of image layer separation as well as stereo matching.  Training our deep net is a simple task, as it does not need to have disparity maps.  Extensive experiments demonstrate the efficacy of our method.
\keywords{stereo computation, image separation, anaglyph, monocular depth estimation, double vision.}
\end{abstract}

\section{Introduction}
Stereo computation (stereo matching) is a well-known and fundamental vision problem, in which a dense depth map $D$ is estimated from two images of the scene from slightly different viewpoints. Typically, one of the cameras is in the left (denoted by $I_L$) and the other in the right (denoted by $I_R$), just like we have left and right two eyes.  Given a single image, it is generally impossible to infer a disparity map, unless using strong semantic-dependent image priors such as those single-image depth-map regression works powered by deep-learning \cite{garg2016unsupervised,monodepth17,Li_2015_CVPR}. Even though these learning based monocular depth estimation methods could predict a reasonable disparity map from a single image, they all assume the input image to be \emph{an original color image}.

In this paper, we propose a novel and original problem, assuming instead one is provided with one \emph{single mixture image} (denoted by $I$) which is a composition of an original stereo image pair $I_L$ and $I_R$, \ie~ $I = f(I_L, I_R)$, and the task is to simultaneously recover both the stereo image pair $I_L$ and $I_R$, and an accurate dense depth-map $D$. Under our problem definition, $f$ denotes different image composition operators that generate the mixture image, which is to be defined in details later. This is a very challenging problem, due to the obvious ill-pose (under-constrained) nature of the task, namely, from one input mixture image $I$ one effectively wants to recover three images ($I_L$, $I_R$, and $D$).

In theory it appears to be a blind signal separation (BSS) task, \ie, separating an image into two different component images. However, conventional methods such as BSS using independent component analysis (ICA) \cite{BSS-ICA} are unsuitable for this problem as they make strong assumptions on the statistical independence between the two components. Under our problem definition, $I_L, I_R$ are highly correlated. In computer vision, image layer separation such as reflection and highlight removal \cite{Layer_Separation_Yang:CVPR-2016,Li2015} are also based on the difference in image statistics, again, unsuitable. Another related topic is image matting \cite{Closed-form-Matting:TPAMI-2008}, which refers to the process of accurate foreground estimation from an image. However it either needs human interaction or depends on the difference between foreground object and background, which cannot be applied to our task.

In this paper, we advocate a novel deep-learning based solution to the above task, by using a simple network architecture. We could successfully solve for a stereo pair $L,R$ and a dense depth map $D$ from a single mixture image $I$. Our network consists of an image separation module and a stereo matching module, where the two modules are optimized jointly. Under our framework, the solution of one module benefits the solution of the other module. It is worth-noting that the training of our network does not require ground truth depth maps.

At a first glance, this problem while intrigue, has pure intellectual interest only, not perhaps no practical use. In contrast, we show this is not the case: in this paper, we show how to use it to solve for three very different vision problems: double vision, de-analygphy and even monocular depth estimation.

The requirement for de-anaglyph is still significant. If search on Youtube, there are hundreds if not thousands of thousands of anaglyph videos, where the original stereo images are not necessarily available. Our methods and the previous work \cite{Williem2015}\cite{Joulin2013} enable the recovery of the stereo images and the corresponding disparity map, which will significantly improve the users' \emph{real 3D experience}. As evidenced in the experiments, our proposed method clearly outperforms the existing work with a wide gap. Last but not least, our model could also handle the task of monocular depth estimation and it comes as a surprise to us: Even with one single mixture image, trained on the KITTI benchmark, our method produces the state of the art depth estimation,  with results more better than those traditional two images based methods.

\section{Setup the stage}
\begin{figure}[!ht]
\begin{center}
\subfigure{
\includegraphics[width=0.45\linewidth]{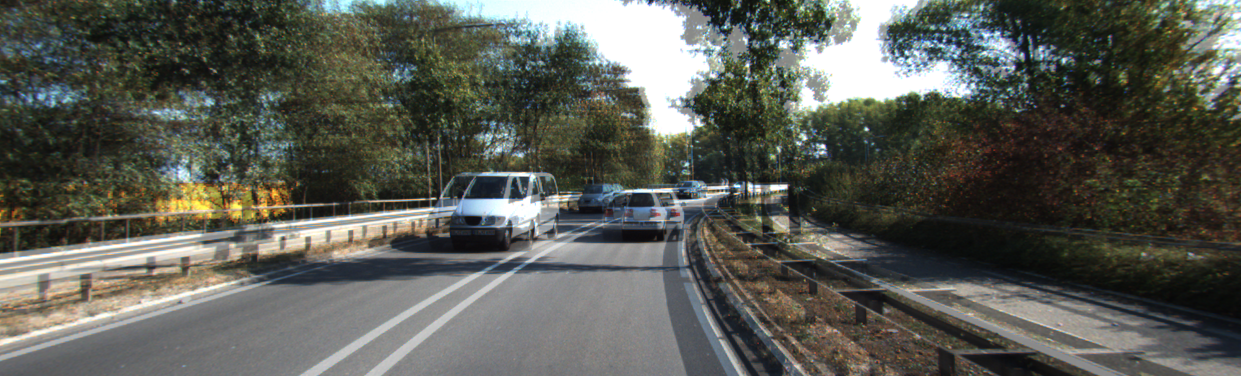}}
\subfigure{
\includegraphics[width=0.45\linewidth]{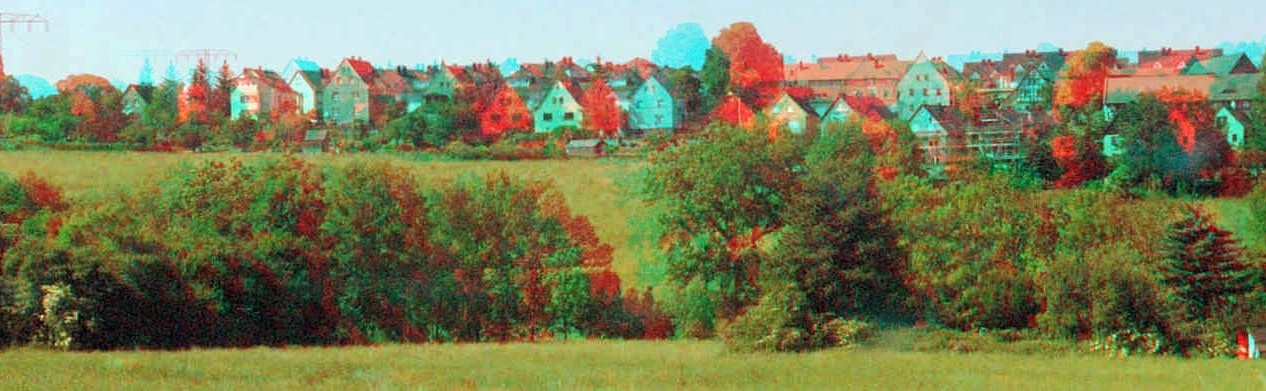}}
\caption{\label{fig:lr} \small Examples of image separation for single image based stereo computation. {\bf Left column:} A double-vision image is displayed here. {\bf Right column:} A red-cyan stereo image contains channels from the left and right images. }
\end{center}
\end{figure}

In this paper we study two special cases of our novel problem of \emph{joint image separation and stereo computation}, namely {\em anaglyph} (red-cyan stereo) and {\em diplopia} (double vision) (see Figure.~\ref{fig:lr}), which have not been well-studied in the past.

\begin{compactenum}[1)]
\item \textbf{Double vision (aka. diplopia)}: {\em Double vision} is the simultaneous perception of two images (a stereo pair) of a single object in the form of a single mixture image. Specifically, under the double vision (diplopia) model (\cf ~Fig.~\ref{fig:lr} (left column)), the perceived image $I = f(I_L, I_R) = (I_L + I_R)/2$, \ie, the image composition is $f$ a direct average of the left and the right images. Note that the above equation shares similarity with the linear additive model in layer separation \cite{Layer_Separation_Yang:CVPR-2016,Layer-Separation-relative-smooth:Li-CVPR-2014,fan2017generic} for reflection removal and raindrop removal, we will discuss the differences in details later.

\item \textbf{Red-Cyan stereo (aka. anaglyph)}: An anaglyph (\cf ~Fig.~\ref{fig:lr} (right column)) is a single image created by selecting chromatically opposite colors (typically red and cyan) from a stereo pair. Thus given a stereo pair $I_L, I_R$, the image composition operator $f$ is defined as $I = f(I_L, I_R)$, where the red channel of $I$ is extracted from the red channel of $I_L$ while its green and blue channels are extracted from $I_R$. De-anaglyph \cite{Williem2015,Joulin2013} aims at estimating both the stereo pair $I_L, I_R$ (color restoration) and computing its disparity maps.
\end{compactenum}

At a first glance, the problem seems impossible as one has to generate two images plus a dense disparity map from \emph{one single input}. However, since the two constitute images are not arbitrary but related by a valid disparity map. Therefore, they must be able to aligned well along the scanlines horizontally. For anaglyph stereo, existing methods \cite{Williem2015,Joulin2013} exploit both image separation constraint and disparity map computation to achieve color restoration and stereo computation. Joulin and Kang \cite{Joulin2013} reconstructed the original stereo pairs given the input anaglyph by using a modified SIFT-flow method \cite{Liu2011}.  Williem \etal ~\cite{Williem2015} presented a method to solve the problem within iterations of color restoration and stereo computation. These works suggest that by properly exploiting the image separation and stereo constraints, it is possible to restore the stereo pair images and compute the disparity map from a single mixture image.

There is little work in computer vision dealing with double vision (diplopia), which is nonetheless an important topic in ophthalmology and visual cognition. The most related works seem to be layer separation \cite{Layer_Separation_Yang:CVPR-2016,Layer-Separation-relative-smooth:Li-CVPR-2014}, where the task is to decompose an input image into two layers corresponding to the background image and the foreground image. However, there are significant differences between our problem and general layer separation. For layer separation, the two layers of the composited image are generally independent and statistically different. In contrast, the two component images are highly correlated for double vision.

Even though there have been remarkable progresses in monocular depth estimation, current state-of-the-art network architectures \cite{garg2016unsupervised,monodepth17} and \cite{Deep3D} cannot be directly applied to our problem. This is because that they depend on a single left/right image input, which is unable to handle image mixture case investigated in this work. Under our problem definition, the two tasks of image separation and stereo computation are tightly  coupled:  stereo computation is not possible without correct image separation; on the other hand, image separation will benefit from disparity computation.

In this paper, we present a unified framework to handle the problem of stereo computation for a single mixture image, which naturally unifies various geometric vision problems such as anaglyph, diplopia and even monocular depth estimation. Our network can be trained with the supervision of stereo pair images only without the need for ground truth disparity maps, which significantly reduces the requirements for training data. Extensive experiments demonstrate that our method achieves superior performances.

\section{Our Method}
In this paper, we propose an end-to-end deep neural network to simultaneously learn image separation and stereo computation from a single mixture image. It can handle a variety forms of problems such as anaglyph, de-diplopia and even monocular depth estimation. Note that existing work designed for either layer-separation or stereo-computation cannot be applied to our problem directly. This is because these two problems are deeply coupled, \ie, the solution of one problem affects the solution of the other problem. By contrast, our formulation to be presented as below, jointly solves both problems.

\begin{figure}[!htp]
\begin{center}
\includegraphics[width=0.8\linewidth]{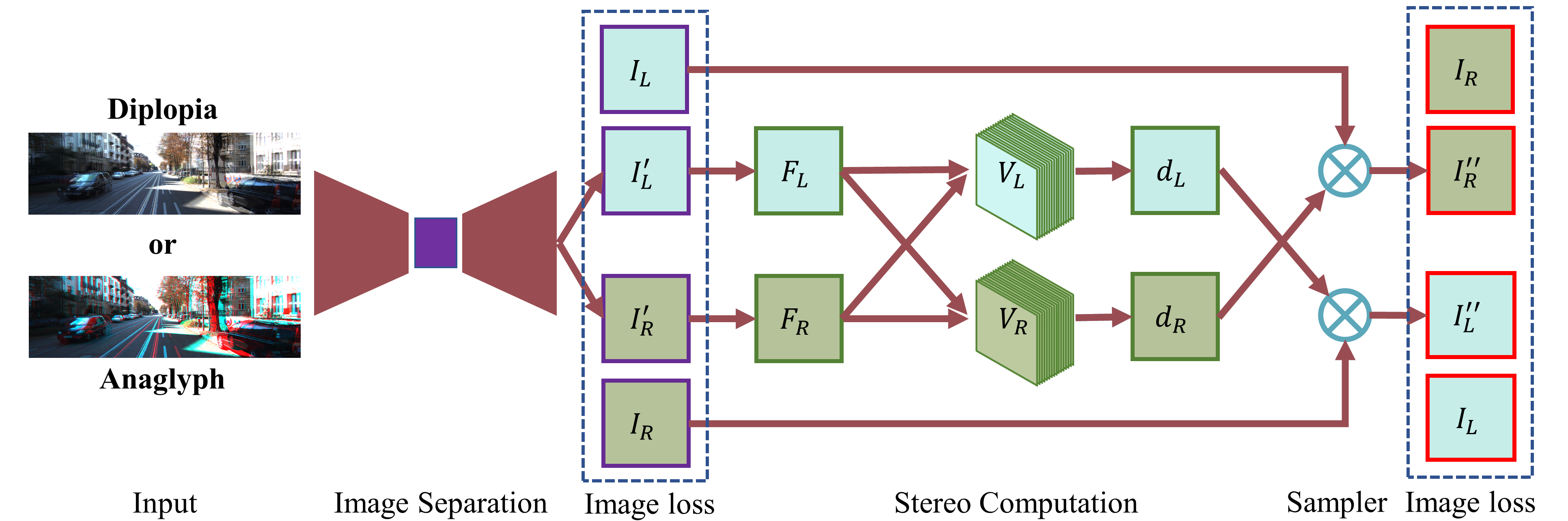}
\caption{\label{fig:rcnet}\small \textbf{Overview of our proposed stereo computation for a single mixture image framework.} Our network consists of an image separation module and a stereo computation module. Take a single mixture image as input, our network simultaneously separates the image into a stereo image pair and computes a dense disparity map.}
\end{center}
\end{figure}

%============================================================
\subsection{Mathematical Formulation}
Under our mixture model, quality of depth map estimation and image separation are evaluated jointly and therefore, the solution of each task can benefit from each other.  Our network model (\cf, Fig.~\ref{fig:rcnet}) consists of two modules, \ie ~an image separation module and a stereo computation module. During network training, only the ground-truth stereo pairs are needed to provide supervisions for both image separation and stereo computation.

By considering both the image separation constraint and the stereo computation constraint in network learning, we define the overall loss function as:
\begin{equation}
\mathcal{L}(\theta_L, \theta_R, \theta_D) = \mathcal{L}_C(\theta_L, \theta_R) + \mathcal{L}_D(\theta_D),
\end{equation}
where $\theta_L, \theta_R, \theta_D$ denote the network parameters corresponding to the image separation module (left image prediction and right image prediction) and the stereo computation module. A joint optimization of $(\theta_L, \theta_R, \theta_D) = \arg\min \mathcal{L}(\theta_L, \theta_R, \theta_D) $ gives both the desired stereo image pair and the disparity map.

%=======================================================
\subsection{Image Separation}

The input single mixture image $I \in \mathbb{R}^{H\times W\times 3}$ encodes the stereo pair image as $I = f(I_L,I_R)$, where $f$ is the image composition operator known a prior. To learn the stereo image pair from the input single mixture image, we present a unified end-to-end network pipeline. Specifically, denote $\mathcal{F}$ as the learned mapping from the mixture image to the predicted left or right image parameterized by $\theta_L$ or $\theta_R$. The objective function of our image separation module is defined as,
\begin{equation}
\alpha_c\mathcal{L}_c(\mathcal{F}(I;\theta_L),I_L) + \alpha_p\mathcal{L}_p (\mathcal{F}(I;\theta_L)),
\label{eq:colorobjl}
\end{equation}
where $I$ is the input single mixture image, $I_L, I_R$ are the ground truth stereo image pair. The loss function $\mathcal{L}$ measures the discrepancy between the predicted stereo images and the ground truth stereo images. The object function for the right image is defined similarly.

In evaluating the discrepancy between images, various loss functions such as $\ell_2$ loss \cite{Iizuka2016}, classification loss \cite{zhang2016colorful} and adversarial loss \cite{pix2pix2016} can be applied. Here, we leverage the pixel-wise $\ell_1$ regression loss as the content loss of our image separation network,
\begin{equation}
\mathcal{L}_c (\mathcal{F}(I;\theta_L), I_L) = \left| \mathcal{F}(I;\theta_L)- I_L\right|.
\label{eq:l1}
\end{equation}
This loss allows us to perform end-to-end learning as compatible with the stereo matching loss and do not need to consider class imbalance problem or add an extra network structure as a discriminator.

Researches on natural image statistics show that a typical real image obeys sparse spatial gradient distributions \cite{Levin2007}. According to Yang \etal ~\cite{Layer_Separation_Yang:CVPR-2016}, such a prior can be represented as the Total Variation (TV) term in energy minimization. Therefore, we have our image prior loss:
\begin{equation}
\mathcal{L}_p (\mathcal{F}(I;\theta_L)) = |\mathcal{F}(I;\theta_L)|_{\rm{TV}} = \left|\nabla \mathcal{F}(I;\theta_L)\right|,
\label{eq:ip}
\end{equation}
where $\nabla$ is the gradient operator.

We design a U-Net architecture \cite{Ronneberger2015} for image separation, which has been used in various conditional generation tasks. Our image separation module consists of 22 convolutional layers. Each convolutional layer contains one convolution-relu pair except for the last layer and we use element-wise add for each skip connection to accelerate the convergence. For the output layer, we utilize a ``tanh'' activation function to map the intensity value between $-1$ and $1$. A detailed description of our network structure is provided in the supplemental material.

The output of our image separation module is a 6 channels image, where the first 3 channels represent the estimated left image $\mathcal{F}(I;\theta_L)$ and the rest 3 channels for the estimated right image $\mathcal{F}(I;\theta_R)$. When the network converges, we could directly use these images as the image separation results. However, for the de-anaglyph task, as there is extra constraint (the mixture happens at channel level), we could leverage the color prior of an anaglyph that the desired image separation (colorization) can be further improved by warping corresponding channels based on the estimated disparity maps.

For the monocular depth estimation task, only the right image will be needed as the left image has been provided as input.

%================================================================
\subsection{Stereo Computation}
The input to the stereo computation module is the separated stereo image pair from the image separation module. The supervision of this module is the ground truth stereo pairs rather than the inputs. The benefit of using ground truth stereo pairs for supervision is that it makes the network not only learn how to find the matching points, but also makes the network to extract features that are robust to the noise from the generated stereo images.

Fig.~\ref{fig:rcnet} shows an overview of our stereo computation architecture, we adopt a similar stereo matching architecture from Zhong \etal ~\cite{SsSMnet2017} without its consistency check module. The benefit for choosing such a structure is that their model can converge within 2000 iterations which makes it possible to train the entire network in an end-to-end fashion. Additionally, removing the need of ground truth disparity maps enables us to access much more accessible stereo images.

Our loss function for stereo computation is defined as:
\begin{equation}
\mathcal{L}_D = \omega_w (\mathcal{L}_w^l+\mathcal{L}_w^r) + \omega_s (\mathcal{L}_s^l+\mathcal{L}_s^r),
\end{equation}
where $\mathcal{L}_w^l, \mathcal{L}_w^r$ denote the image warping appearance loss, $\mathcal{L}_s^l, \mathcal{L}_s^r$ express the smoothness constraint on the disparity map.

Similar to $\mathcal{L}_c$, we form a loss in evaluating the image similarity by computing the pixel-wise $\ell_1$ distance between images. We also add a structural similarity term SSIM \cite{SSIM2004} to improve the robustness against illumination changes across images. The appearance loss $\mathcal{L}^l_w$ is derived as:
\begin{equation}
\mathcal{L}_w^l (I_L, I_L^{''}) = \frac{1}{N}\sum\lambda_1\frac{1-\mathcal{S}(I_L, I_L^{''})}{2} + \lambda_2\left| I_L- I_L^{''}\right|,
\end{equation}
where $N$ is the total number of pixels and $I_L^{''}$ is the reconstructed left image. $\lambda_1, \lambda_2$ balance between structural similarity and image appearance difference. According to \cite{monodepth17}, $I_L^{''}$ can be fully differentially reconstructed from the right image $I_R$ and the right disparity map $d_R$ through bilinear sampling \cite{STN2015}.

For the smoothness term, similar to \cite{monodepth17}, we leverage the Total Variation (TV) and weight it with image's gradients. Our smoothness loss for disparity field is:
\begin{equation}
\mathcal{L}_s^l = \frac{1}{N}\sum\left| \nabla_u d_L\right| e^{-\left| \nabla_u I_L\right|}+ \left| \nabla_v d_L\right| e^{-\left| \nabla_v I_L\right|}.
\end{equation}

%====================================================================

\subsection{Implementation Details}
We implement our network in TensorFlow \cite{tensorflow} with 17.1M trainable parameters. Our network can be trained from scratch in an end-to-end fashion with a supervision of stereo pairs and optimized using RMSProp \cite{Tieleman2012} with an initial learning rate of $1\times 10^{-4}$. Input images are normalized with pixel intensities level ranging from -1 to 1. For the KITTI dataset, the input images are randomly cropped to $256\times 512$, while for the Middlebury dataset, we use $384\times 384$. We set disparity level to 96 for the stereo computation module. For weighting loss components, we use $\alpha_c = 1, \alpha_p = 0.2, \omega_w = 1, \omega_s = 0.05$. We set $\lambda_1 = 0.85,\lambda_2 = 0.15$ throughout our experiments.  Due to the hardware limitation (Nvidia Titan Xp), we only use batch size 1 during network training.

%================================================================
\section{Experiments and Results}
In this section, we validate our proposed method and present experimental evaluation for both de-anaglyph and de-diplopia (double vision). For experiments on anaglyph images, given a pair of stereo images, the corresponding anaglyph image can be generated by combining the red channel of the left image and the green/blue channels of the right image. Any stereo pairs can be used to quantitatively evaluate the performance of de-anaglyph. However, since we also need to quantitatively evaluate the performance of anaglyph stereo matching, we use two stereo matching benchmarking datasets for evaluation: Middlebury dataset \cite{Scharstein2002} and KITTI stereo 2015 \cite{Menze2015CVPR}. Our network is initially trained on the KITTI Raw dataset with 29000 stereo pairs that listed by \cite{monodepth17} and further fine-tuned on Middlebury dataset. To highlight the generalization ability of our network, we also perform qualitative experiments on random images from Internet. For de-diplopia (double vision), we synthesize our inputs by averaging stereo pairs. Qualitative and quantitative results are reported on KITTI stereo 2015 benchmark \cite{Menze2015CVPR} as well. Similar to the de-anaglyph experiment, we train our initial model on the KITTI raw dataset.

%==============================================================
\subsection{Advantages of joint optimization}
Our framework consists of image separation and stereo computation, where the solution of one subtask benefits the solution of the other subtask. Direct stereo computation is impossible for a single mixture image. To analyze the advantage of joint optimization, we perform ablation study in image separation without stereo computation and the results are reported in Table~\ref{tab:abl}. Through joint optimization, the average PSNR increases from $19.5009$ to $20.0914$, which demonstrates the benefit of introducing the stereo matching loss in image separation.

\begin{table}[!htp]
\small
\centering
\tabcolsep=0.39cm
\begin{tabular}{|c|c|c|}
\hline
Metric & Image separation only & Joint optimization \\ \hline
PSNR & 19.5009 &20.0914  \\ \hline
\end{tabular}
\caption{\label{tab:abl}\small {Ablation study of image separation on KITTI.}}
\end{table}

\subsection{Evaluation of Anaglyph Stereo}
We compare the performance of our method with two state-of-the-art de-anaglyph methods: Joulin \etal ~\cite{Joulin2013} and Williem \etal ~\cite{Williem2015}. Evaluations are performed on two subtasks: stereo computation and image separation (color restoration).

\begin{figure}[H]
\begin{center}
\subfigure{
\includegraphics[height=0.15\linewidth]{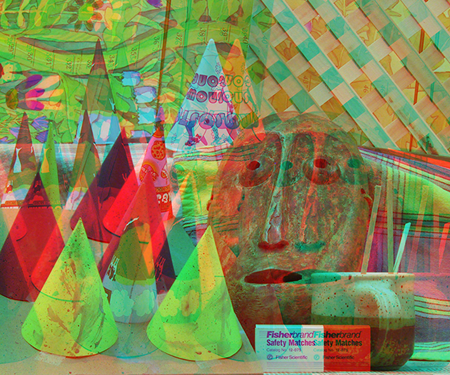} }
\subfigure{
\includegraphics[height=0.15\linewidth]{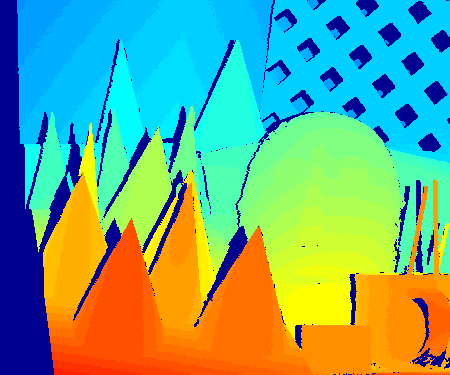} }
\subfigure{
\includegraphics[height=0.15\linewidth]{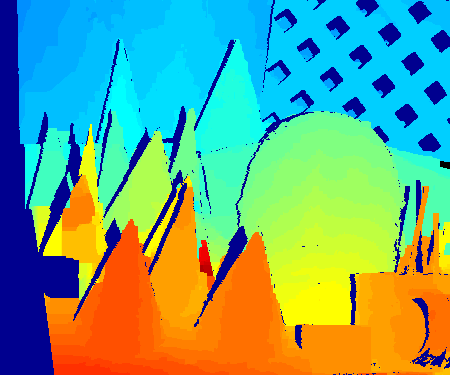} }
\subfigure{
\includegraphics[height=0.15\linewidth]{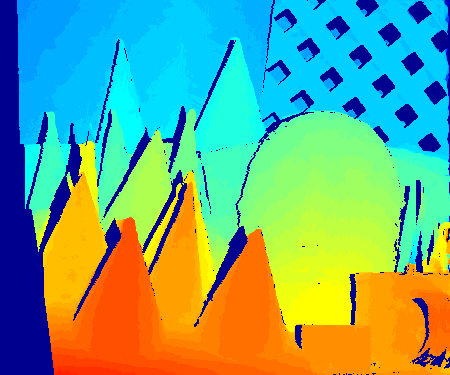} }
\caption{\label{fig:mba} \small {Qualitative stereo computation results on the Middlebury dataset by our method.} From left to right: input anaglyph image, ground truth disparity map, disparity map generated by Williem \etal ~\cite{Williem2015} and our method.}
\end{center}
\end{figure}

\begin{figure}[h]
\begin{center}
\includegraphics[width=0.42\linewidth]{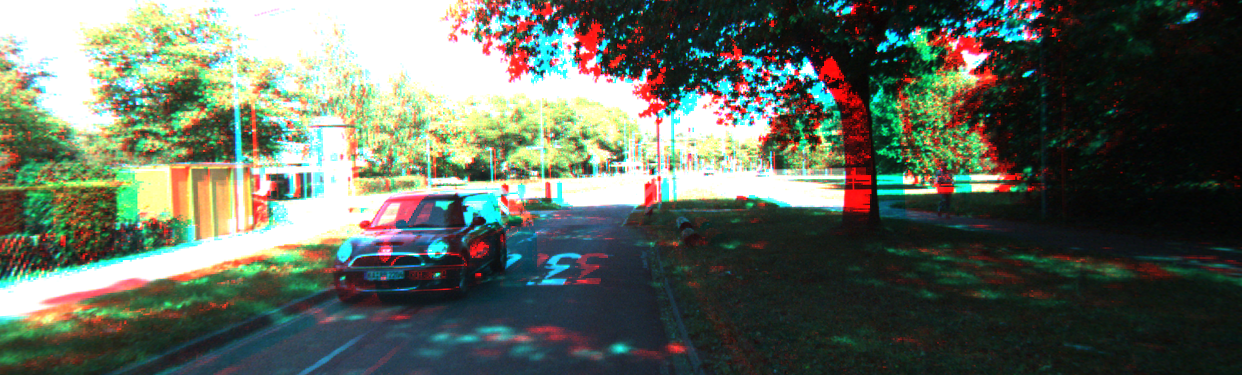}
\includegraphics[width=0.42\linewidth]{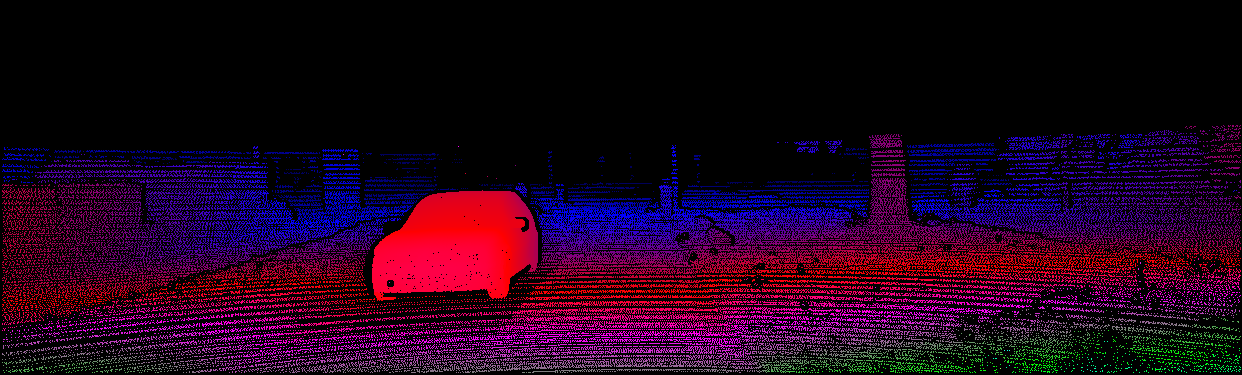}\\
\includegraphics[width=0.42\linewidth]{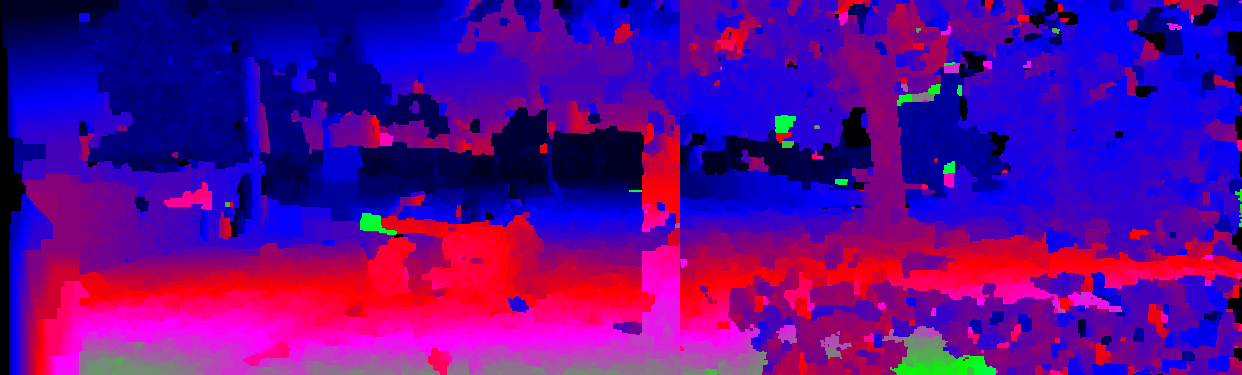}
\includegraphics[width=0.42\linewidth]{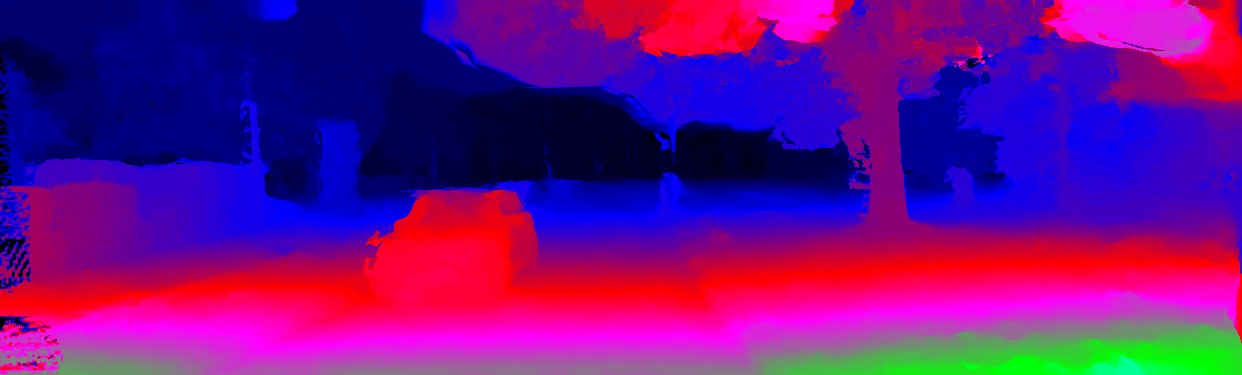}
\caption{\label{fig:kitti15d} \small {Qualitative disparity map recovery results on KITTI-2015 of our method.} Top row: Input anaglyph image and ground truth disparity map. Bottom row: Result of Williem \etal ~\cite{Williem2015} and our result.}
\end{center}
\end{figure}

\noindent
\textbf{Stereo Computation.}
We present qualitative comparison of estimated disparity maps in Fig.~\ref{fig:mba} for Middlebury \cite{Scharstein2002} and in Fig.~\ref{fig:kitti15d} for KITTI 2015 \cite{Menze2015CVPR}. Stereo pairs in Middlebury are indoor scenes with multiple handcrafted layouts and the ground truth disparities are captured by highly accurate structural light sensors. On the other hand, the KITTI stereo 2015 consists of 200 outdoor frames in their training set, which is more challenging than the Middlebury dataset. The ground truth disparity maps are generated by sparse LIDAR points and CAD models.

\begin{figure}[H]
\begin{center}
\includegraphics[width=0.50\linewidth]{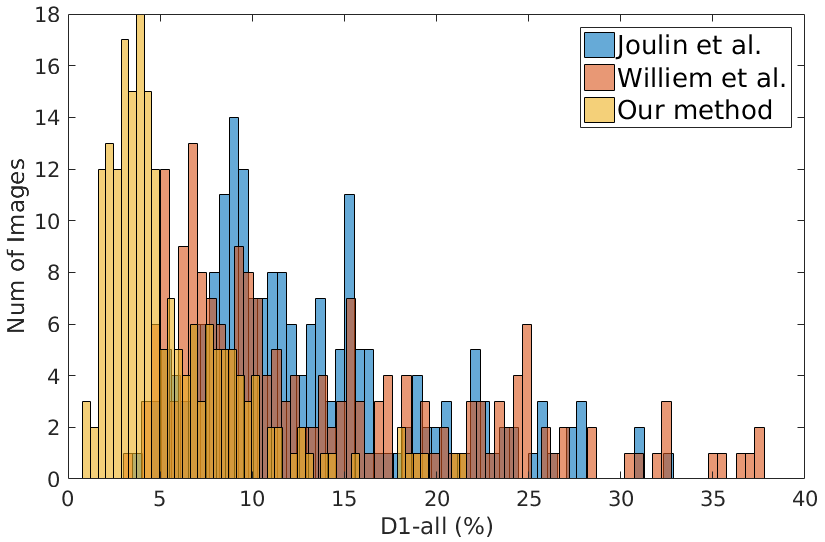}
\caption{\label{fig:d1_allkitticy}{\small Disparity map estimation results comparison on the KITTI stereo 2015 dataset.}}
\end{center}
\end{figure}

On both datasets, our method can generate more accurate disparity maps than previous ones from visual inspection. It can be further evidenced by the quantitative results of bad pixel percentage that shown in Table.~\ref{tab:mb} and Fig.~\ref{fig:d1_allkitticy}. For the Middlebury dataset, our method achieves $32.55\%$ performance leap than Williem \etal~ \cite{Williem2015} and $352.28\%$ performance leap than Joulin \etal ~\cite{Joulin2013}. This is reasonable as Joulin \etal ~\cite{Joulin2013} did not add disparity into its optimization. For the KITTI dataset, we achieve an average bad pixel ratio (denoted as D1\_all) of $5.96\%$ with 3 pixel thresholding across 200 images in the training set as opposed to $13.66\%$ by Joulin \etal ~\cite{Joulin2013} and $14.40\%$ by Williem \etal ~\cite{Williem2015}.

\begin{table}[H]
\centering
\tabcolsep=0.136cm
\begin{tabular}{|c | c | c | c | c | c |}
\hline
{Method} & {Tsukuba} & {Venus} & {Cones} & { Teddy} & { Mean}\\ \hline
Joulin \cite{Joulin2013}  & 14.02 & 25.90 & 23.49 & 43.85 & 26.82 \\ \hline
Williem \cite{Williem2015} & 12.53 & 2.24 & 8.00 & 8.68 & 7.86 \\ \hline
Ours  &  {\bf 10.44} & {\bf 1.25} & {\bf 4.51} & {\bf 7.51} & {\bf 5.93}\\ \hline
\end{tabular}
\caption{\label{tab:mb}\small {Performance comparison in disparity map estimation for de-anaglyph on the Middlebury dataset. We report the bad pixel ratio with a threshold of 1 pixel. Disparities are scaled according to the provided scaling factor on the Middlebury dataset.}}
\end{table}

\noindent
\textbf{Image Separation.}
As an anaglyph image is generated by concatenating the red channel from the left image and the green and blue channels from the right image, the original color can be found by warping the corresponding channels based on the estimated disparity maps. We leverage such a prior for de-anaglyph and adopt the post-processing step from Joulin \etal ~\cite{Joulin2013} to handle occluded regions. Qualitative and quantitative comparison of image separation performance are conducted on the Middlebury and KITTI datasets. We employ the Peak Signal-to-Noise Ratio (PSNR) to measure the image restoration quality.

\begin{figure}[!htp]
\begin{center}
\subfigure{
\includegraphics[width=0.47\linewidth]{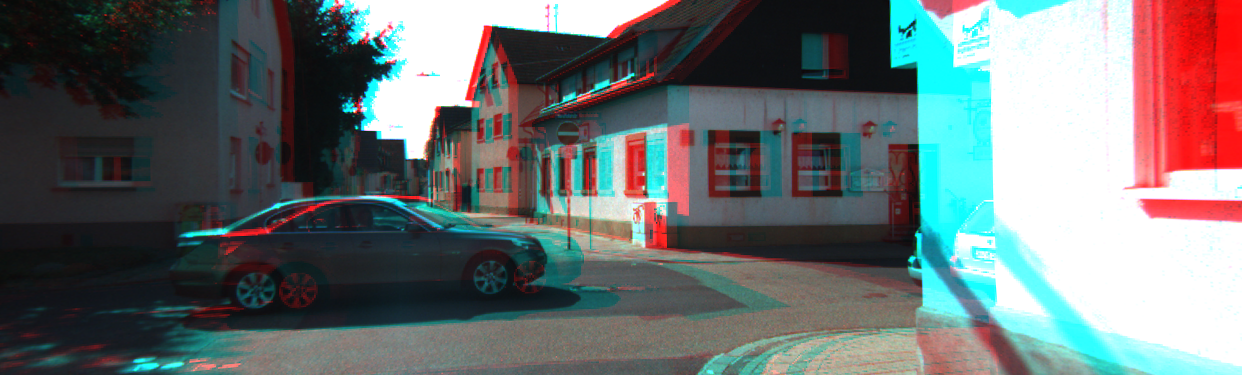} }
\subfigure{
\includegraphics[width=0.47\linewidth]{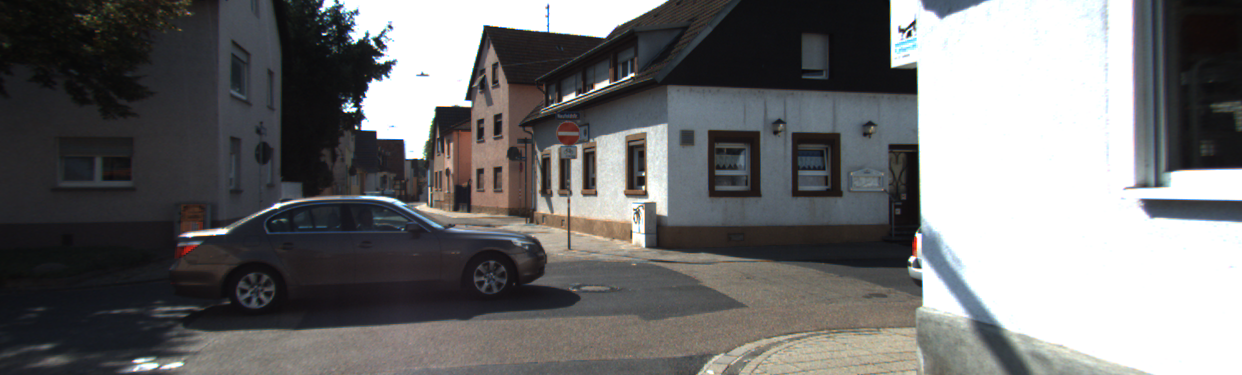} }
\subfigure{
\includegraphics[width=0.47\linewidth]{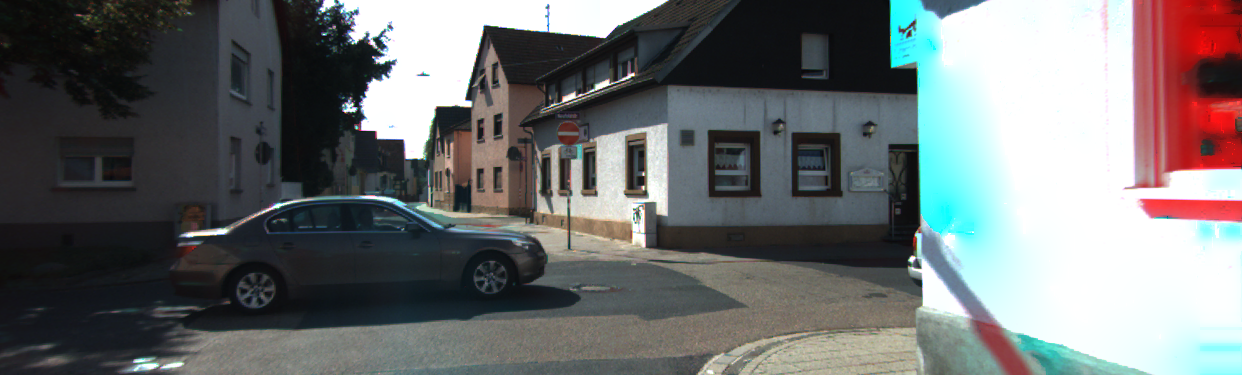} }
\subfigure{
\includegraphics[width=0.47\linewidth]{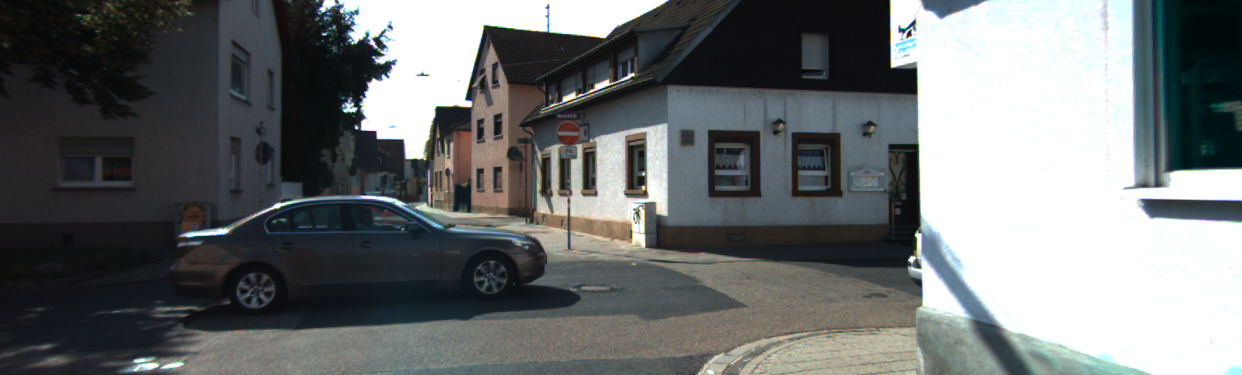} }
\caption{\label{fig:kitti15}\small {Qualitative image separation results on the KITTI-2015 dataset.} Top to bottom: Input, ground truth, result from Williem \etal ~\cite{Williem2015}, our result. Our method successfully recovers the correct color  of the large textureless region on the right of the image while the other method fails.}
\end{center}
\end{figure}

\begin{figure}[!htp]
\begin{center}
\subfigure{
\includegraphics[height=0.25\linewidth]{middlebury/2.png} }
\subfigure{
\includegraphics[height=0.25\linewidth]{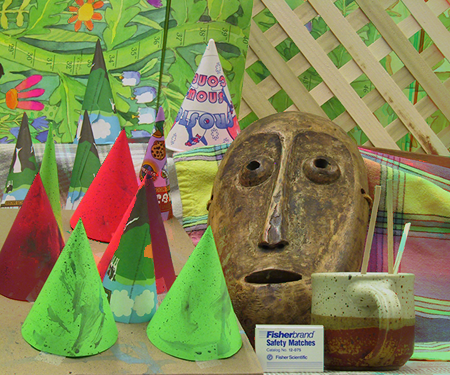} }
\subfigure{
\includegraphics[height=0.25\linewidth]{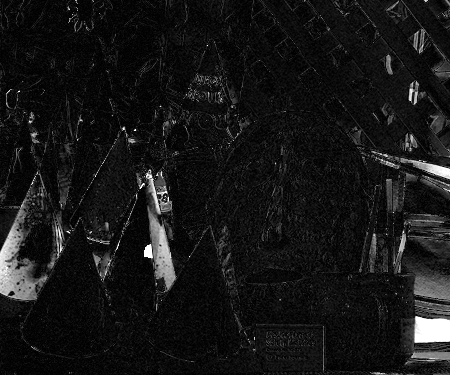} }
\subfigure{
\includegraphics[height=0.25\linewidth]{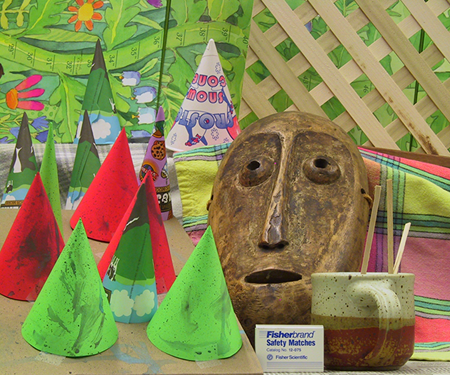} }
\subfigure{
\includegraphics[height=0.25\linewidth]{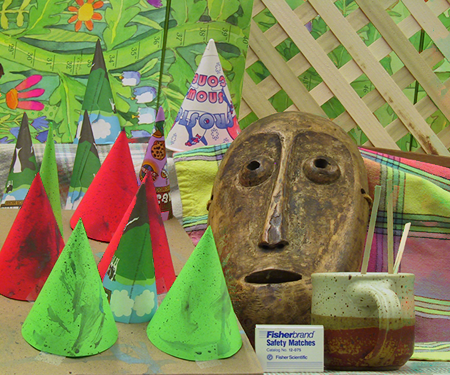} }
\subfigure{
\includegraphics[height=0.25\linewidth]{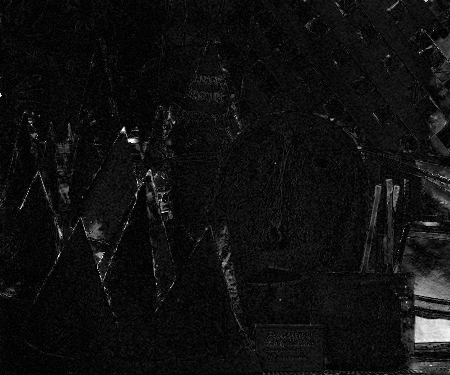} }
\caption{\label{fig:mb}\small {Qualitative comparison in image separation (restoration) on the Middlebury dataset.} The first column shows the input anaglyph image and the ground truth image. The results of Williem \etal ~\cite{Williem2015} (top) and our method (bottom) with their corresponding error maps are shown in the second and the third column.}
\end{center}
\end{figure}

Qualitative results for both datasets are provided in Fig.~\ref{fig:kitti15} and Fig.~\ref{fig:mb}. Our method is able to recover colors in the regions where ambiguous colorization options exist as those areas rely more on the correspondence estimation, while other methods tend to fail in this case.

Table~\ref{tab:mbc} and Table~\ref{tab:kitti_c} report the performance comparison between our method and state-of-the-art de-anaglyph colorization methods: Joulin \etal ~\cite{Joulin2013} and Williem \etal ~\cite{Williem2015} on the Middlebury dataset and on the KITTI dataset correspondingly. For the KITTI dataset, we calculated the mean PSNR throughout the total 200 images of the training set. Our method outperforms others with a notable margin. Joulin \etal ~\cite{Joulin2013} is able to recover relatively good restoration results when the disparity level is small, such as Tsukuba, Venus, and KITTI. When the disparity level doubled, its performance drops quickly as for Cone and Teddy images. Different with Williem \etal ~\cite{Williem2015}, which can only generate disparity maps at pixel level, our method is able to further optimize the disparity map to sub-pixel level, therefore achieves superior performance in both stereo computation and image restoration (separation).

\begin{table}[!htp]
\small
\centering
\tabcolsep=0.035cm
\begin{tabular}{|c|c|c|c|c|c|c|c|c|}
\hline
\multirow{2}{*}{Method} & \multicolumn{2}{c|}{Tsukuba} & \multicolumn{2}{c|}{Venus} & \multicolumn{2}{c|}{Cones} & \multicolumn{2}{c|}{Teddy} \\ \cline{2-9}
                        & Left          & Right        & Left         & Right       & Left         & Right       & Left         & Right       \\ \hline
Joulin \cite{Joulin2013}                   & 30.83         & 32.88        & 29.66        & 31.97       & 21.52        & 24.54       & 21.16        & 24.59       \\ \hline
Williem \cite{Williem2015}                   & 30.32         & 31.12        & 31.41        & 34.93       & 23.68        & 26.17       & 23.14        & 31.05      \\ \hline
Ours                & {\bf 32.99}         & {\bf 35.31 }       & {\bf 35.05}        & {\bf 37.74}       & {\bf 26.31 }       & {\bf 30.17}       & {\bf 28.44}        & {\bf 35.53 }      \\ \hline
\end{tabular}
\caption{\label{tab:mbc} \small {Performance comparisons (PSNR) in image separation (restoration) for the task of de-anaglyph on the Middlebury dataset.}}
\end{table}

\begin{table}[!htp]
\centering
\tabcolsep=0.22cm
\begin{tabular}{|c|c|c|c|c|}
\hline
Dataset                & View  & Joulin \cite{Joulin2013} & Williem \cite{Williem2015}  & Ours \\ \hline
\multirow{2}{*}{KITTI} & Left  & 25.13 &　24.57         & {\bf 26.30 }        \\ \cline{2-5}
                       & Right & 27.19 &  26.94          & {\bf 28.76 }        \\ \hline
\end{tabular}
\caption{\label{tab:kitti_c} \small {Performance comparisons (PSNR) in image separation (restoration) for the task of de-anaglyph on the KITTI dataset.}}
\end{table}

\begin{figure}[!htp]
\begin{center}
\subfigure{
\includegraphics[width=0.28\linewidth,height = 1.5cm]{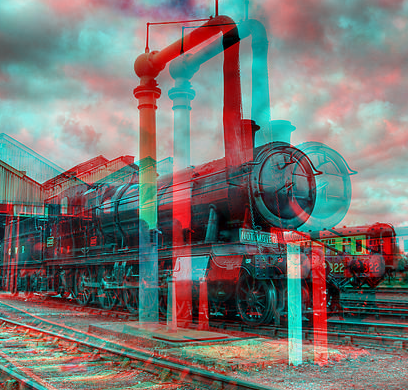} }
\subfigure{
\includegraphics[width=0.28\linewidth,height = 1.5cm]{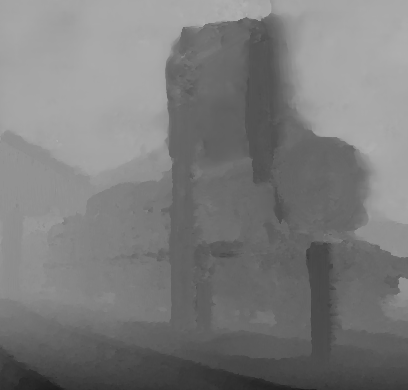} }
\subfigure{
\includegraphics[width=0.28\linewidth,height = 1.5cm]{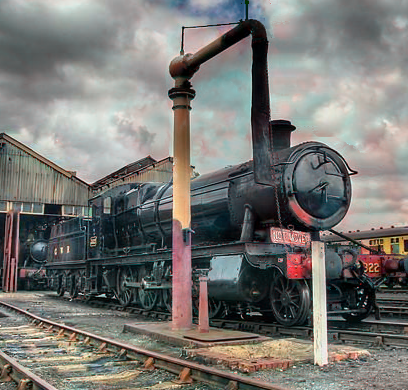} }
\subfigure{
\includegraphics[width=0.28\linewidth,height = 1.5cm]{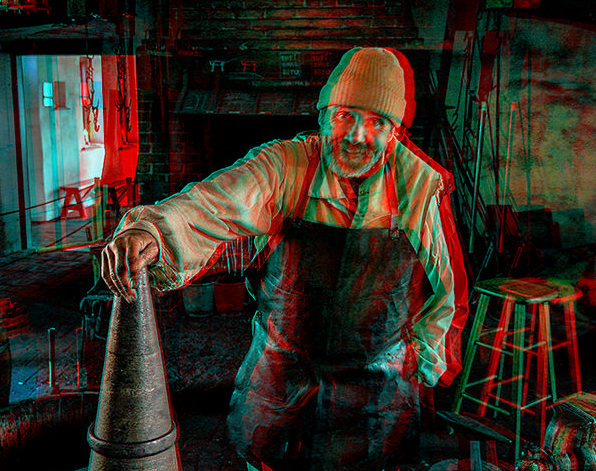} }
\subfigure{
\includegraphics[width=0.28\linewidth,height = 1.5cm]{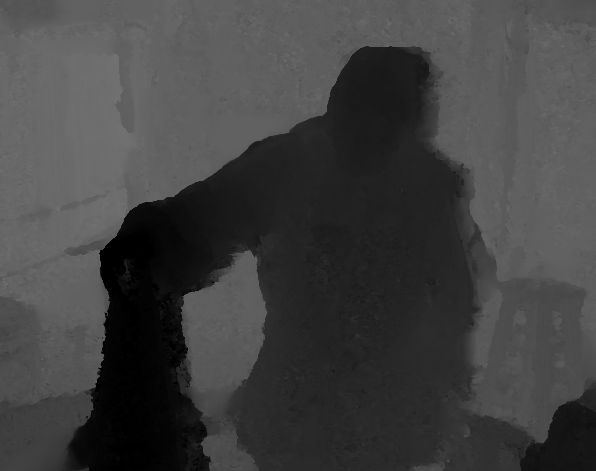} }
\subfigure{
\includegraphics[width=0.28\linewidth,height = 1.5cm]{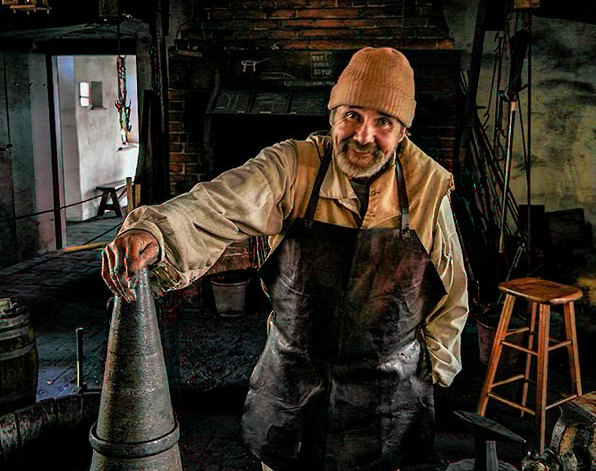} }
\subfigure{
\includegraphics[width=0.28\linewidth,height = 1.5cm]{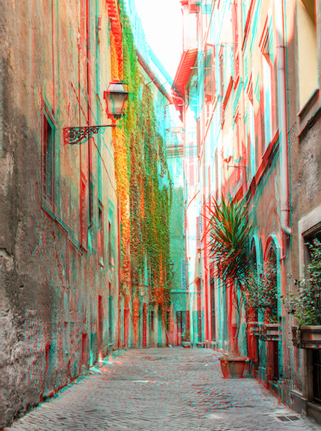} }
\subfigure{
\includegraphics[width=0.28\linewidth,height = 1.5cm]{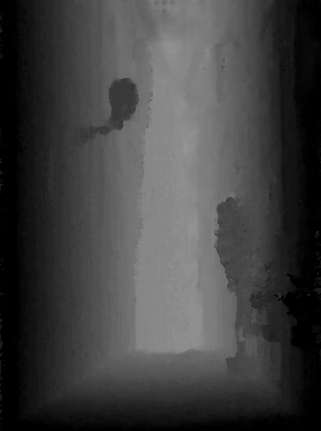} }
\subfigure{
\includegraphics[width=0.28\linewidth,height = 1.5cm]{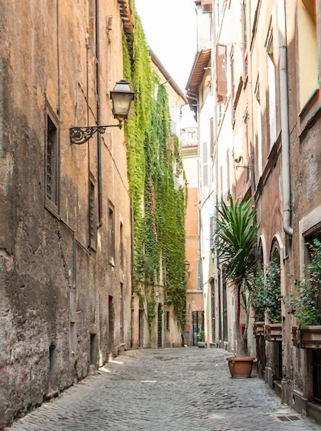} }
\subfigure{
\includegraphics[width=0.28\linewidth,height = 1.5cm]{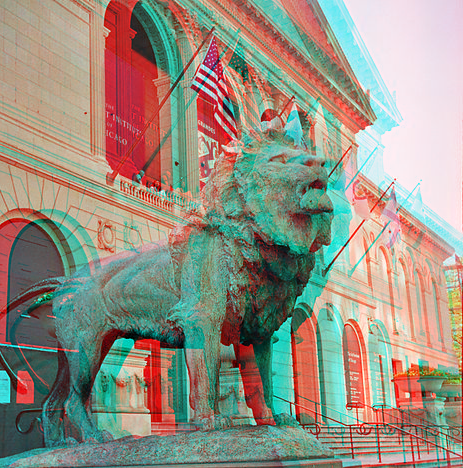} }
\subfigure{
\includegraphics[width=0.28\linewidth,height = 1.5cm]{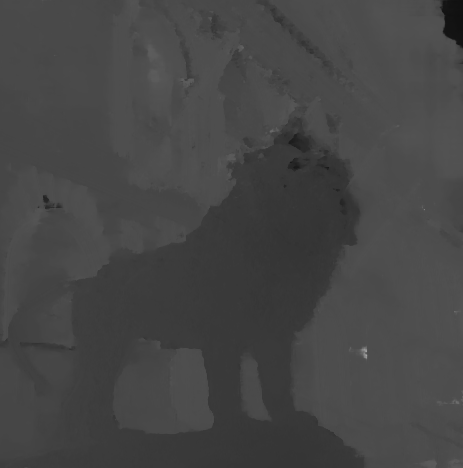} }
\subfigure{
\includegraphics[width=0.28\linewidth,height = 1.5cm]{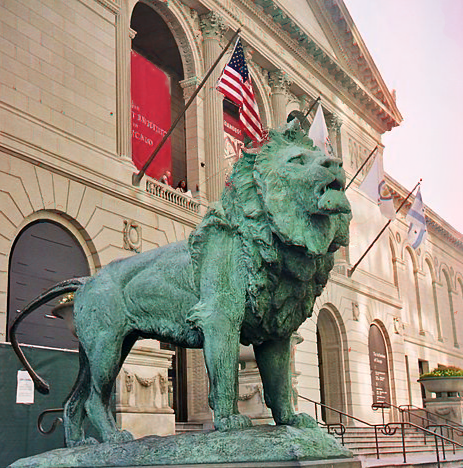} }
\caption{\label{fig:wild}\small {Qualitative stereo computation and image separation results for real world anaglyph images downloaded from the Internet.} Left to right: input anaglyph images, disparity maps of our method, image separation results of our method.}
\end{center}
\end{figure}

\noindent
\textbf{Anaglyph in the wild.}
One of the advantages of conventional methods is their generalization capability. They can be easily adapt to different scenarios with or without parameter changes. Deep learning based methods, on the other hand, are more likely to have a bias on specific dataset. In this section, we provide qualitative evaluation of our method on anaglyph images downloaded from the Internet to illustrate the generalization capability of our method. Our method, even though trained on the KITTI dataset which is quite different from all these images, achieves reliable image separation results as demonstrated in Fig.~\ref{fig:wild}. This further confirms the generalization ability of our network model.

%===============================================================
\subsection{Evaluation for double-vision unmixing}
\begin{figure}[!htp]
\begin{center}
\includegraphics[width=0.50\linewidth]{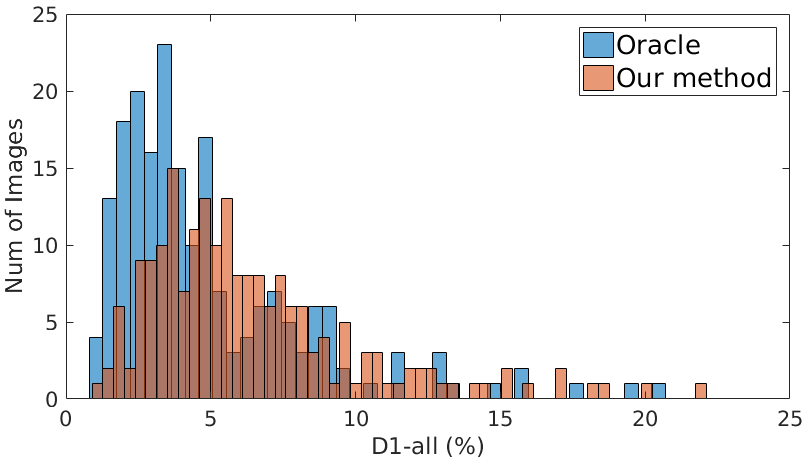}
\caption{\label{fig:d1_allkittidiplo}\small {Stereo computation results on the KITTI 2015 dataset. The oracle disparity map is computed with clean stereo images.}}
\end{center}
\end{figure}

\begin{figure}[h]
\begin{center}
\subfigure{
\includegraphics[width=0.47\linewidth]{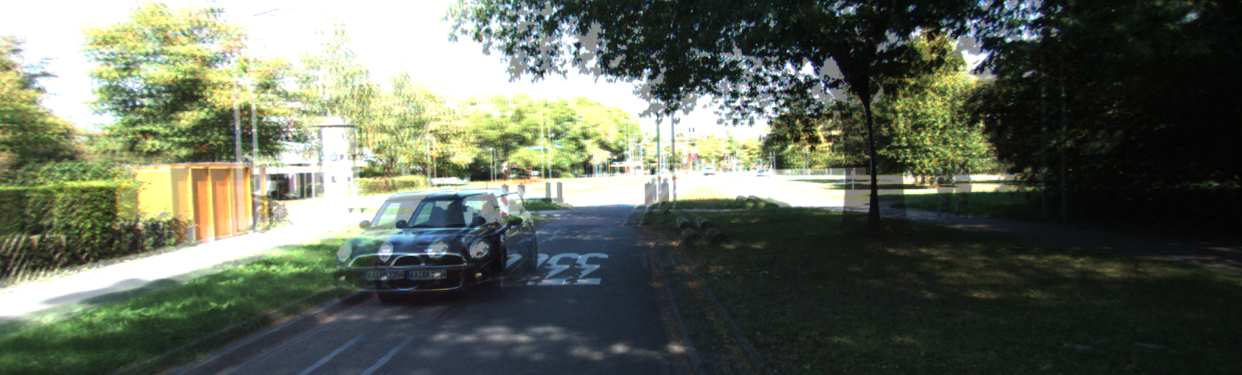} }
\subfigure{
\includegraphics[width=0.47\linewidth]{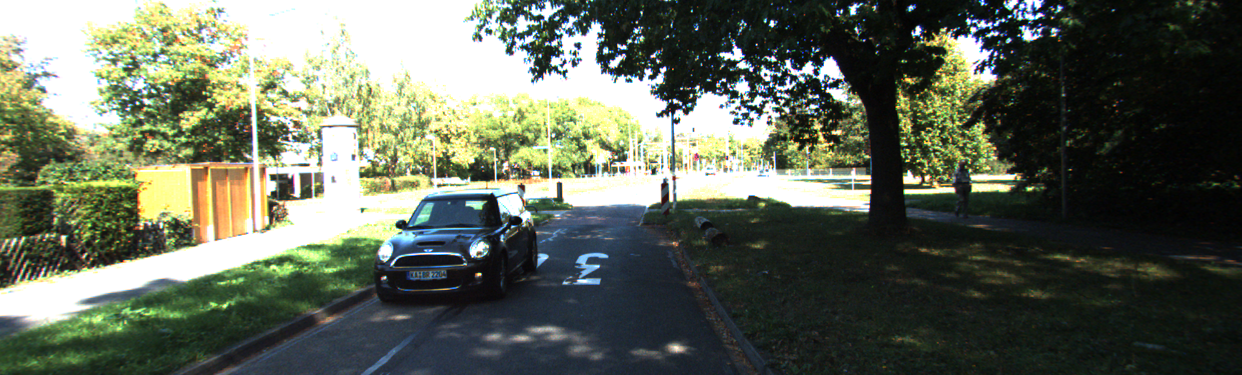} }
\subfigure{
\includegraphics[width=0.47\linewidth]{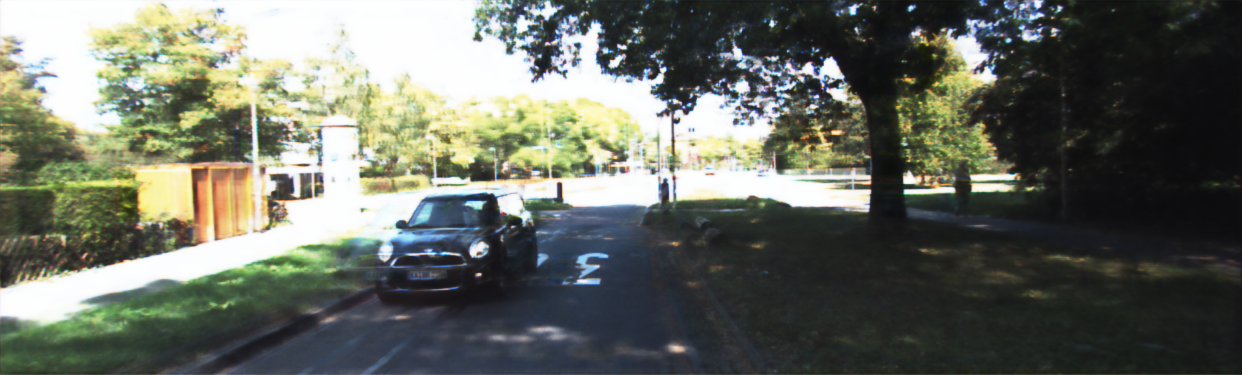} }
\subfigure{
\includegraphics[width=0.47\linewidth]{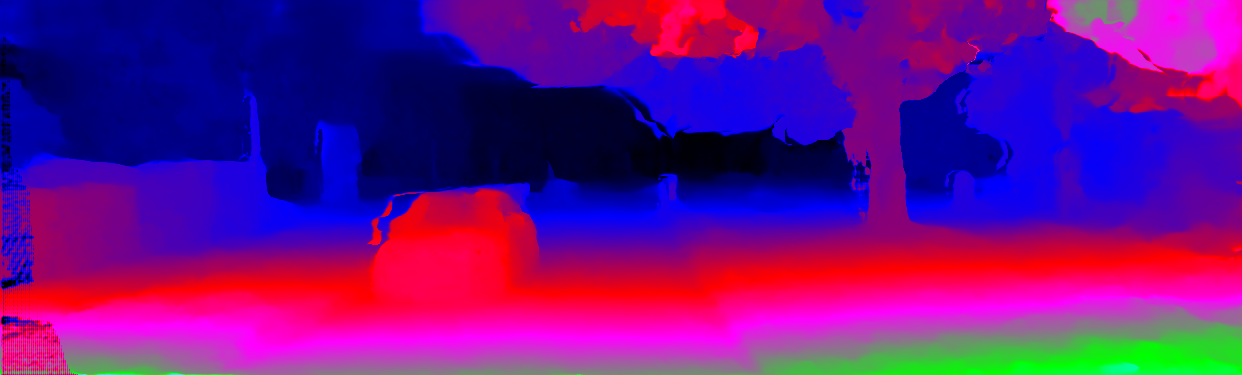} }
\caption{\label{fig:kitti15lr}\small {Qualitative diplopia unmixing results by our proposed method.} Top to bottom,left to right: input diplopia image, ground truth left image, restored left image by our method, and our estimated disparity map.}
\end{center}
\end{figure}

Here, we evaluate our proposed method for unmixing of double-vision image,  where the input image is the average of a stereo pair.  Similar to anaglyph, we evaluate our performance based on the estimated disparities and reconstructed stereo pair on the KITTI stereo 2015 dataset. For disparity evaluation, we use the oracle disparity maps (that are computed with clean stereo pairs) as a reference in Fig.~\ref{fig:d1_allkittidiplo}. The mean bad pixel ratio of our method is $6.67\%$, which is comparable with the oracle's performance as $5.28\%$. For image separation, we take a layer separation method \cite{Li2014} as a reference. A quantitative comparison is shown in Table~\ref{tab:kitti_c_lr}. Conventional layer separation methods tend to fail in this scenario as the statistic difference between the two mixed images is minor which violates the assumption of these methods. Qualitative results of our method are shown in Fig.~\ref{fig:kitti15lr}.

\begin{table}[!htp]
\centering
\tabcolsep=0.473cm
\begin{tabular}{|c|c|c|c|}
\hline
Dataset                & View  & Li \etal \cite{Li2014} & Ours \\ \hline
\multirow{2}{*}{KITTI} & Left  & 17.03  & {\bf 24.38 }        \\ \cline{2-4}
                       & Right & 7.67   & {\bf 24.55 }        \\ \hline
\end{tabular}
\caption{\label{tab:kitti_c_lr}\small {Performance comparison in term of PSNR for the task of image restoration from double-vision images on the KITTI dataset.}}
\end{table}

%=========================================================================
\section{Beyond Anaglyph and Double-Vision}
Our problem definition also covers the problem of monocular depth estimation, which aims at estimating a depth map from a single image \cite{Eigen2014,monodepth17,libo_2018,Li_2015_CVPR}. Under this setup, the image composition operator $f$ is defined as $I = f(I_L, I_R) = I_L$ or $I = f(I_L, I_R) = I_R$, \ie, the mixture image is the left image or the right image. Thus, monocular depth estimation is a special case of our problem definition.

\begin{figure*}[!htp]
\begin{center}
\subfigure{
\includegraphics[width=0.295\linewidth]{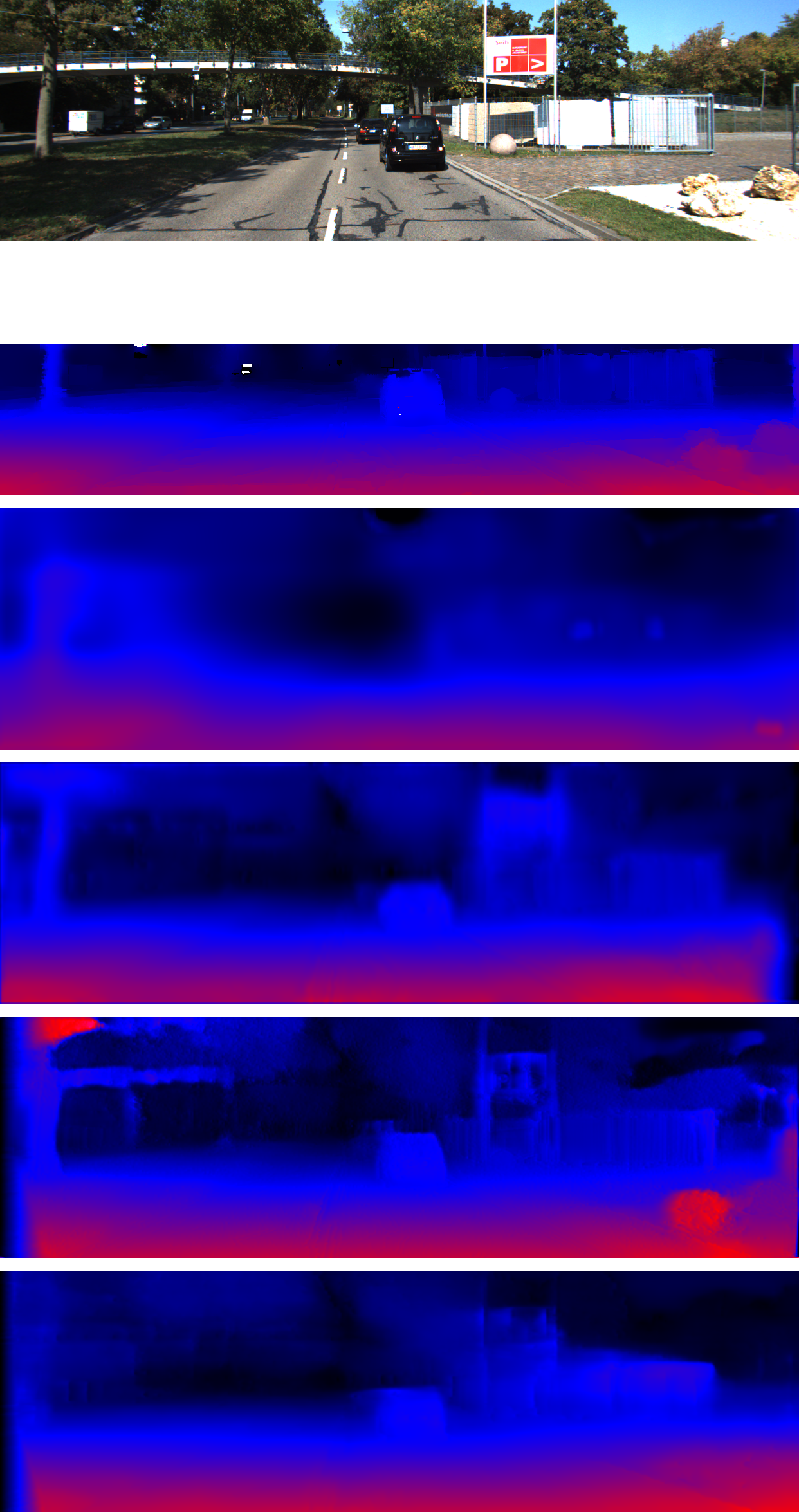} }
\subfigure{
\includegraphics[width=0.295\linewidth]{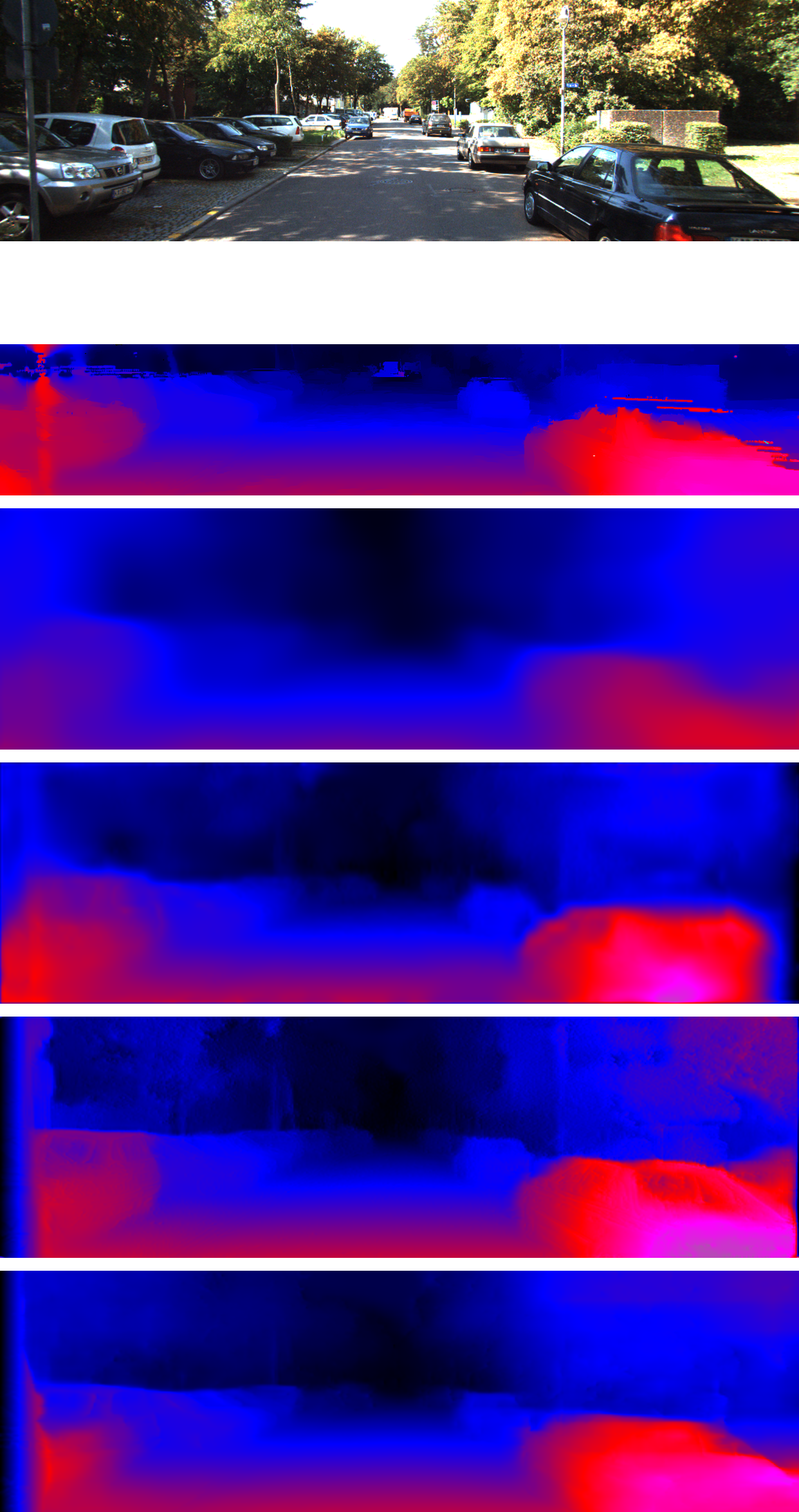} }
\subfigure{
\includegraphics[width=0.295\linewidth]{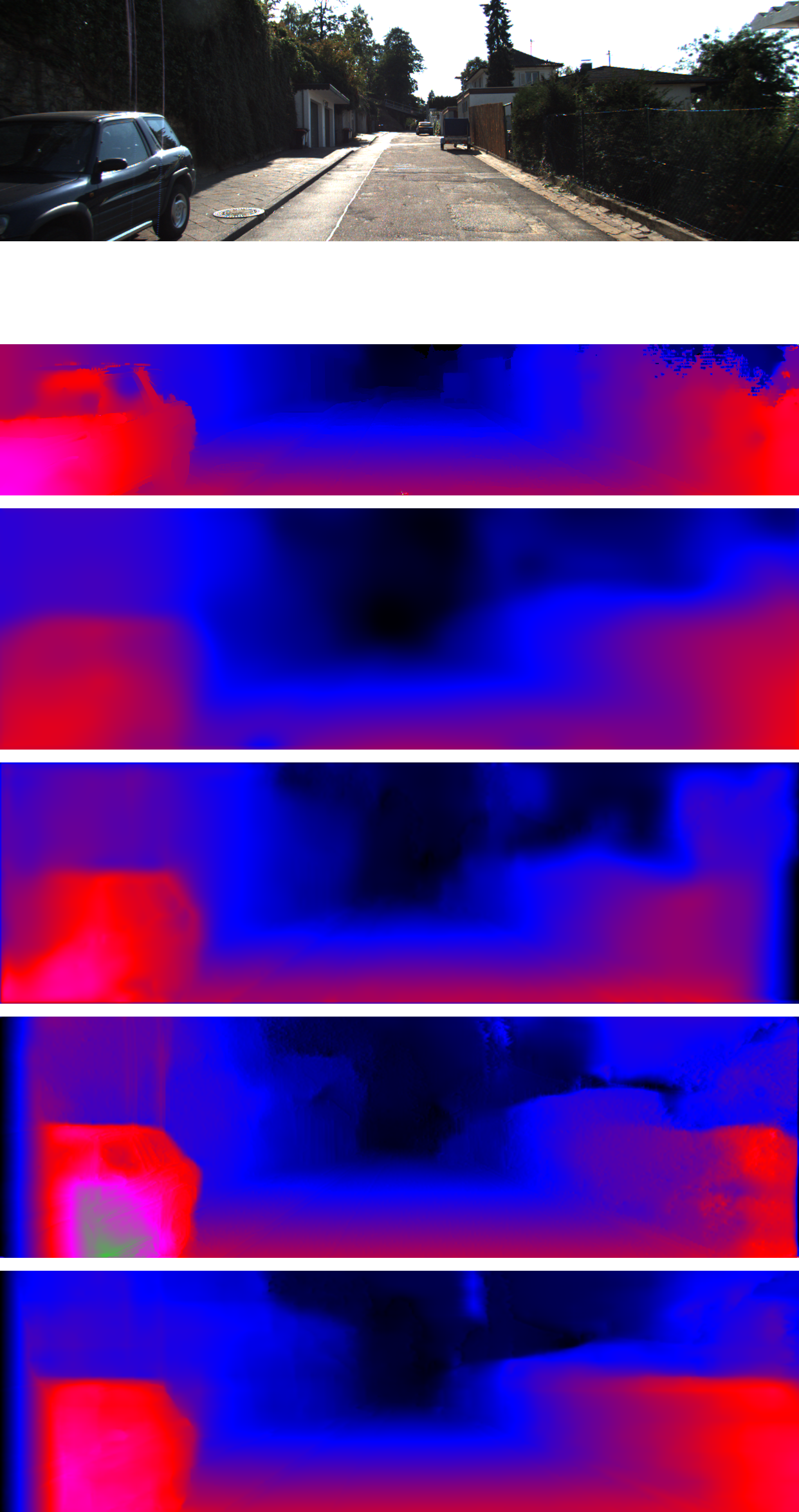} }
\caption{\label{fig:kitti154} \small Qualitative monocular estimation evaluations on the KITTI-2015 dataset: Top to bottom: left image, ground truth, results from Zhou \etal \cite{zhou2017unsupervised}, results from Garg \etal \cite{garg2016unsupervised}, results from Godard \etal \cite{monodepth17} and our results. Since the ground truth depth points are very sparse, we interpolated it with a color guided depth painting method \cite{YangAR2014} for better visualization.}
\end{center}
\end{figure*}

We evaluated our framework for monocular depth estimation on the KITTI 2015 dataset. Quantitative results and qualitative results are provided in Table~\ref{tab:kitti_mono} and Fig.~\ref{fig:kitti154}, where we compare our method with state-of-the-art methods \cite{garg2016unsupervised}, \cite{zhou2017unsupervised} and \cite{monodepth17}. Our method, even designed for a much more general problem, outperforms both \cite{garg2016unsupervised} and \cite{zhou2017unsupervised} and achieves quite comparable results with \cite{monodepth17}.

\begin{table*}[h]
\centering
\small
\tabcolsep=0.045cm
\begin{tabular}{|c|c|c|c|c|c|c|c|}
\hline
Methods               & Abs Rel  & Sq Rel & RMSE &RMSE log & $\delta<1.25$ & $\delta<1.25^2$ & $\delta<1.25^3$ \\ \hline
Zhou \etal \cite{zhou2017unsupervised}  &0.183	&1.595	&6.709	&0.270	&0.734	&0.902	&0.959\\
Garg \etal \cite{garg2016unsupervised}   & 0.169   & 1.080 & 5.104 &0.273 & 0.740 & 0.904 & 0.962 \\
Godard \etal \cite{monodepth17}   & 0.140 &0.976 &4.471 &0.232 &0.818 &0.931 &0.969 \\ \hline
Ours   &  0.126 &0.835 & 3.971  &0.207 & 0.845 & 0.942 &  0.975 \\ \hline
\end{tabular}
\caption{\label{tab:kitti_mono} \small Monocular depth estimation results on the KITTI 2015 dataset using the split of Eigen \etal ~\cite{Eigen2014}. Our model is trained on 22,600 stereo pairs from the KITTI raw dataset listed by \cite{monodepth17} by 10 epochs. Depth metrics are from Eigen \etal ~\cite{Eigen2014}. Our performance is better than the state-of-the-art method \cite{monodepth17}.}
\end{table*}

\section{Conclusion}
This paper has defined a novel problem of stereo computation from a single mixture image, where the goal is to separate a single mixture image into a pair of stereo images--from which a legitimate disparity map can be estimated. This problem definition naturally unifies a family of challenging and practical problems such as anaglyph, diplopia and monocular depth estimation. The problem goes beyond the scope of conventional image separation and stereo computation. We have presented a deep convolutional neural network based framework that jointly optimizes the image separation module and the stereo computation module. It is worth noting that we do not need ground truth disparity maps in network learning. In the future, we will explore additional problem setups such as ``alpha-matting''. Other issues such as occlusion handling and extension to handle video should also be considered.

\noindent
\textbf{Acknowledgements}
Y. Zhong's PhD scholarship is funded by Data61. Y. Dai is supported in part by National 1000 Young Talents Plan of China, Natural Science Foundation of China (61420106007, 61671387), and ARC grant (DE140100180). H. Li's work is funded in part by ACRV (CE140100016). The authors are very grateful to NVIDIA's generous gift of GPUs to ANU used in this research.

\bibliographystyle{splncs}
\bibliography{Deep_Anaglyphs_Reference}

%\includepdf[pages={1,2,3,4,5}]{632_supp.pdf}

\end{document}